# BioConceptVec: creating and evaluating literature-based biomedical concept embeddings on a large scale

Qingyu Chen, Kyubum Lee, Shankai Yan, Sun Kim, Chih-Hsuan Wei, Zhiyong Lu[*]

National Center for Biotechnology Information (NCBI), National Library of Medicine (NLM),

National Institutes of Health (NIH), Bethesda, Maryland, United States of America

*Corresponding author

Email: zhiyong.lu@nih.gov

# Abstract

A massive number of biological entities, such as genes and mutations, are mentioned in the biomedical literature. The capturing of the semantic relatedness of biological entities is vital to many biological applications, such as protein-protein interaction prediction and literature-based discovery. Concept embeddings—which involve the learning of vector representations of concepts using machine learning models—have been employed to capture the semantics of concepts. To develop concept embeddings, named-entity recognition (NER) tools are first used to identify and normalize concepts from the literature, and then different machine learning models are used to train the embeddings. Despite multiple attempts, existing biomedical concept embeddings generally suffer from suboptimal NER tools, small-scale evaluation, and limited availability.

In response, we employed high-performance machine learning-based NER tools for concept recognition and trained our concept embeddings, BioConceptVec, via four different machine learning models on ~30 million PubMed abstracts. BioConceptVec covers over 400,000 biomedical concepts mentioned in the literature and is of the largest among the publicly available biomedical concept embeddings to date. To evaluate the validity and utility of BioConceptVec, we respectively performed two intrinsic evaluations (identifying related concepts based on drug-gene and gene-gene interactions) and two extrinsic evaluations (protein-protein interaction prediction and drug-drug interaction extraction), collectively using over 25 million instances from nine independent datasets (17 million instances from six intrinsic evaluation tasks and 8 million instances from three extrinsic evaluation tasks), which is, by far, the most comprehensive to our best knowledge. The intrinsic evaluation results demonstrate that BioConceptVec consistently has, by a large margin, better performance than existing concept embeddings in identifying similar and related concepts. More importantly, the extrinsic evaluation results demonstrate that using BioConceptVec with advanced deep learning models can significantly improve performance in downstream bioinformatics studies and biomedical text-mining applications.

Our BioConceptVec embeddings and benchmarking datasets are publicly available at

https://github.com/ncbi-nlp/BioConceptVec.

# Author Summary

Capturing the semantics of related biological concepts, such as genes and mutations, is of significant importance to many research tasks in computational biology such as protein-protein interaction detection, gene-drug association prediction, and biomedical literature-based discovery. Here, we propose to leverage state-of-the-art text mining tools and machine learning models to learn the semantics via vector representations (aka. embeddings) of over 400,000 biological concepts mentioned in the entire PubMed abstracts. Our learned embeddings, namely BioConceptVec, can capture related concepts based on their surrounding contextual information in the literature, which is beyond exact term match or co-occurrence-based methods. BioConceptVec has been thoroughly evaluated in multiple bioinformatics tasks consisting of over 25 million instances from nine different biological datasets. The evaluation results demonstrate that BioConceptVec has better performance than existing methods in all tasks. Finally, BioConceptVec is made freely available to the research community and general public.

# Introduction

In the biomedical domain, one primary application of text mining is to extract knowledge within the biomedical literature automatically [1]. Specifically, identifying important concepts (mentioned in the literature, such as gene/proteins, diseases, and mutations, is critical to biocuration [2], literature-based knowledge discovery [3], and many downstream applications [4-6]. Previous studies have used different words such as concepts, entities, names, and mentions to refer to the same topic in the biomedical domain. Here, we use *bio-concepts* for consistency. Similar to the use of word embeddings, capturing the representation of bio-concepts plays a vital role in biomedical applications such as biomedical relation extraction [12] and document classification [13]. Existing studies use the term *concept embeddings*, which is a special kind of word embedding [7-9]. According to the literature, a concept embedding may contain only the concept vectors [10], or it may contain vectors of both concepts and common words [11]. Named entity recognition (NER) tools or concept dictionaries are often used to identify and normalize concepts in a consistent format [10].

Since 2014, word embedding models have revolutionized how to represent text. In these models, each word is represented as a high dimensional vector [12, 13]. The vector representations are learned on large-scale free text corpora via unsupervised learning. Primary methods include training the embeddings based on (1) averaged surrounding context words, such as the continuous bag-of-words (cbow) model in word2vec [14], (2) weighted context words, such as the skip-gram model in word2vec, (3) global co-occurrence statistics, such as GloVe [15], and (4) word n-grams, such as fastText [16]. The use of vector representations can capture related words from different lexicons, such as cancer and tumor. This overcomes the limitations of traditional bag-of-words approaches that rely on exact term matching [17]. To date, text-mining applications have rapidly adopted word embeddings. For instance, the use of embeddings have shown promising performance in biomedical applications such as biomedical document classification [18], sentence retrieval [19], and question answering [20].

It is known that biomedical concepts have a high degree of ambiguity [21]. The same words can be used to describe different types of concepts in free text; for example, AP2 can be the name of a gene (https://www.ncbi.nlm.nih.gov/gene/?term=2167), a chemical (https://meshb.nlm.nih.gov/record/ui?ui=C417523), or a cell-line (https://web.expasy.org/cellosaurus/CVCL_1147). Conversely, the same concepts can have different names; for example, the HER2 gene has at least 10 different synonyms mentioned in text (https://www.ncbi.nlm.nih.gov/gene/2064).  In addition, a bio-concept can span multiple words; for example, *serum and glucocorticoid-induced protein kinase* is the name of a gene (SGK1, https://www.ncbi.nlm.nih.gov/gene/6446). Therefore, accurate NER is essential prior to training concept embeddings.

We present a detailed summary of the existing bio-concept embeddings in Table 1. These studies have used various corpora (mainly electronic health records (EHR), combined with medical claims, biomedical corpora, or Wikipedia) and several training methods (mainly word2vec, while some used GloVe and fastText) to train concept embeddings. Overall, the primary method paradigm is consistent among these studies and generally involves two steps. In the first step, NER tools are applied to identify and normalize target concepts and to replace the mentions in the text as a preprocessing to the corpora. In the second (embedded training) step, embedding training occurs, whereby standard word embedding training methods, such as word2vec, are employed. Note that we consider concept embeddings trained on knowledge bases, such as gene2vec [22], as different work because knowledge bases are distinct from free-text collections. For example, knowledge bases contain concepts already curated either manually or semi-automatically; therefore, training concept embeddings via knowledge bases does not require NER tools. In addition, the relationships between concepts in knowledge bases already have been organized in a structured format, such as ontologies. Free text, however, is unstructured, and training embeddings on free text occurs purely in an unsupervised way. Also note that individual knowledge bases contain only specific types of concepts by design. By contrast, a wide spectrum of concept types are described in the literature.

Despite these recent efforts, past studies share some limitations. As shown in Table 1, existing studies used NER tools to recognize and normalize Unified Medical Language System (UMLS) concepts [23]. A long series of evaluation studies demonstrate that the effectiveness of these NER tools fluctuates dramatically for different types of UMLS concepts [24-28]. For example, Hassanzadeh et al. evaluated the NER tools used by the studies in Table 1 and found that the F1-score ranged from 5% to 75% for different types of UMLS concepts [24]. Likewise, Reátegui et al. found that the F1-score of the NER tools varied from 44% to 96% for different types of diseases [26]. Importantly, errors produced in the NER step may diminish the effectiveness of bio-concept embeddings. For example, low precision, such as a non-concept word wrongly identified as a bio-concept by NER tools, will bias the context or nearby words of the true bio-concepts when training embeddings. Similarly, low recall, such as true bio-concepts that are not identified by NER tools, will reduce the number of training instances and decrease the concept coverage of bio-concept embeddings.

Second, almost no studies had evaluated the effectiveness of concept embeddings in extrinsic evaluations. The evaluation of word embeddings can be broadly categorized into two types (i.e., intrinsic and extrinsic) [29]. Intrinsic evaluations are commonly accomplished via an unsupervised setting or using weakly supervised labels, whereas extrinsic evaluations are often performed via a supervised setting in downstream applications. As shown in Table 1, only one study [8] performed extrinsic evaluations for heart failure, predicting whether a patient would be diagnosed as having heart failure based on the associated clinical notes. The study used a basic long short-term memory (LSTM) model with randomly initialized embedding as the baseline and replaced the randomly initialized embedding with the proposed concept embedding to compare the performance. Although the results demonstrated that the proposed concept embedding has better performance, the study (1) did not compare the results with those of other existing concept embeddings and (2) did not compare the results with those of the state-of-the-art model that had achieved the highest performance on that task [30].

Further, importantly, the existing concept embeddings are designed primarily for concepts and applications in the clinical domain, whereas concept embeddings for the biological domain remain to be developed. As shown in Table 1, existing studies used UMLS concepts and mainly used EHR data as the training corpora. Correspondingly, the evaluation focuses on clinical applications, i.e., the evaluation datasets are generated from EHR data. For example, most of the studies evaluated the two datasets, UMNSRS (Medical Residents Relatedness Set)-Similarity [31] and UMNSRS-Relatedness [31], each consisting of ~600 pairs of clinical concepts derived from EHR data and annotated by physicians. Similarly, the above extrinsic evaluation of heart-failure prediction is also based on a patient's clinical notes [8]. Developing embeddings for biological concepts and applications is also important.

In response, we propose BioConceptVec, a collection of concept embeddings on primary biological concepts mentioned in the biomedical literature. Fig 1 shows an overview of our study. Specifically, the study has three primary contributions:

1. To our knowledge, we are the first study to use machine learning-based NER tools to recognize and normalize biological concepts for training bio-concept embeddings. Specifically, we employed PubTator, a state-of-the-art NER system with concept annotations for the entire PubMed abstracts [32]. It contains over 400,000 concepts, which is the largest among the publicly available concept embeddings. For example, our evaluation of the human gene coverage shows that BioConceptVec covers 33% more gene concepts than the existing concept embeddings.

2. We conducted large-scale intrinsic and extrinsic evaluations to quantify the validity and utility of BioConceptVec. The intrinsic evaluations contain ~18 million instances from six datasets. BioConceptVec has significantly higher performance (up to 10% improvement) than the existing concept embeddings and is consistent across multiple datasets. The extrinsic evaluations cover two downstream applications: protein-protein interaction (PPI) prediction, consisting of ~8 million PPIs from the STRING database [33], and drug-drug interaction (DDI) classification, consisting of ~5,000 DDIs from a community-recognized gold standard dataset. The extrinsic evaluation results demonstrate that the deep learning models that use BioConceptVec can significantly improve the state-of-the-art performance, achieving an AUC of 0.95 for predicting PPIs and an F1-score of 0.80 for extracting DDIs.

3. We make all of the embeddings and evaluation datasets publicly available. The embeddings and datasets can be downloaded via https://github.com/ncbi-nlp/BioConceptVec. We also provide a Jupyter notebook that contains code examples for users to get started.

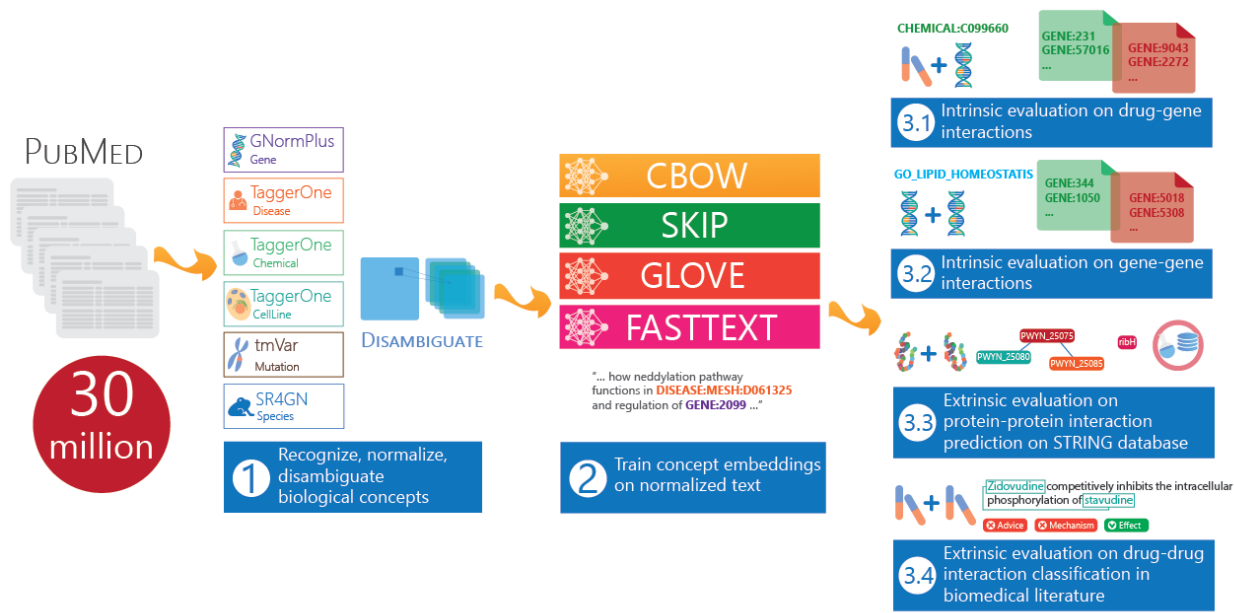

Fig 1. An overview of our study. BioConceptVec was trained on PubMed abstracts, which consists of ~30 million documents. (1) We employed PubTator, which contains four NER tools, to annotate and normalize the concepts. (2) We trained four concept embeddings on the normalized corpus. (3) We conducted both intrinsic evaluations on drug-gene interactions and gene-gene interactions, and extrinsic evaluations on protein-protein interaction prediction and drug-drug interaction extraction to evaluate the effectiveness of BioConceptVec.

# Materials and Methods

## Training corpus and method

### NER step: using PubTator to annotate biological concepts

We trained concept embeddings on the ~30 million abstracts in the entire PubMed. We followed the preprocessing pipeline from [34] (the code is publicly available via https://github.com/ncbi-nlp/BioSentVec). As noted, the first step of bio-concept embedding development is to use NER tools to identify the target concept mentions (e.g., "estrogen receptor") and to further normalize the mentions to the concept identifiers (e.g., "NCBI Gene: 2099"). As an example, shown in Fig 2, a targeted concept (i.e., MLN4924) is identified and normalized to a chemical concept: MESH:C539933. Due to the requirement of high-quality concept normalization for the concept embeddings, we applied PubTator to annotate the full PubMed abstracts. PubTator [32] is a PubMed-scale resource that utilizes four NER tools (i.e., TaggerOne [35], GNormPlus [36], tmVar [37], and SR4GN [38]) with a recent deep learning-based module for disambiguating conflict mentions [39] (when the mentions are annotated by two or more concept taggers) to recognize six key biological concepts (i.e., genes, mutations, diseases, chemicals, cell lines, and species). Table in S1 Table provides a summary of the state-of-the-art performance of the NER tools in PubTator on various benchmarking datasets.

Fig 2. Identified bio-concept in-text and the normalized versions (one instance per type) in PubTator.

## Embedding training step: using word2vec, GloVe, and fastText to produce BioConceptVec

We trained concept embeddings on the full collection of PubMed abstracts after concept recognition via PubTator, i.e., identified named entities are replaced with bio-entity types and IDs (e.g., Disease_MESH_D008288) before training. To our knowledge, there is no agreement on which embedding model is the most effective in biomedical domains. For example, Wang et al. [40] showed that fastText achieved the highest performance in biomedical event trigger detection versus other word embeddings [40], whereas Jin et al. [41] found that word2vec has better performance in biomedical sentence classification [41]. In this study, we therefore trained four different word embeddings, cbow, skip-gram, GloVe, and fastText such that future studies can choose our concept embeddings according to their specific requirements.

In general, the methods to train word embeddings can be categorized into two groups: window-based and matrix factorization-based [15]. The major distinction between these two categories is that window-based methods aim to learn the semantics of words based on local context, i.e., words within a pre-defined window size, whereas matrix factorization-based methods aim to learn the semantics of words based on global statistics of words in corpora. word2vec and fastText belong to the first category while GloVe belongs to the second category. word2vec has two versions: cbow, training a model using context words as input to predict a target word, and skip-gram: reversely using a target word to predict context words [14]. fastText is an extension of word2vec, using character n-grams to represent a word [42]. In contrast, GloVe is dramatically different from word2vec and fastText. It builds a matrix based on global co-occurrences between the words and then applies matrix factorization.

As mentioned, fastText represents each word as a set of character n-grams. In the case of bio-concept embeddings, however, each bio-concept should be considered a unit. Thus, when training with fastText, we disabled the n-grams representation for bio-concepts (in contrast, for the words that are not bio-concepts, we still used the default n-grams representation in fastText).

The values of hyperparameters for training embeddings are summarized in Table 2. Our choice of hyperparameters is based on similar studies in the past and other related work in the general domain.

## Hyperparameters and other methods for comparison

To directly compare with the existing concept embeddings, we used the exact hyperparameter values from Yu et al. [43] as the default setting. As shown in Table 1, of the three publicly available concept embeddings, it is the only concept embedding trained on PubMed. The other two were trained on EHR data. We measured the concept overlap in terms of genes and found that concept embeddings trained on EHR data contain a significantly fewer number of genes than do embeddings trained on PubMed. Thus, we did not compare with those two EHR-driven methods.

Yu et al. [43] used cbow to train the concept embeddings and their hyperparameters are summarized in Table 2. Hence, under the same parameter settings, we firstly trained a common cbow word embedding on PubMed abstracts, as a baseline. Common word embeddings do not contain vectors for normalized bio-concepts. The words in a bio-concept name, however, often exist in common word embeddings. For example, the TOR3A gene (https://www.ncbi.nlm.nih.gov/gene/64222) does not exist in a common word embedding, but the words of its name *torsin family 3 member A* all exist. Thus, we averaged the word vectors based on the bio-concept name to represent the concept vector. Averaged vectors are used as a strong baseline for many embedding-related tasks, such as sentence similarity [44] and sentiment analysis [45]. We refer to the averaged word embedding baseline as BioAvgWord (cbow). As such, we are able to directly compare BioConceptVec (cbow) with the two baselines: BioAvgWord (cbow) and the concept embedding provided by Yu et al.

In addition, we trained and assessed BioConceptVec (cbow) under different parameters but keeping the same values for minimal word occurrences (so that embeddings share the same vocabulary), learning rate and training epochs (so that embeddings share the same optimization procedure). For each of the other hyperparameters, we selected two representative values that were used in the previous studies on embeddings [46, 47], as shown in Table 2 (other values). Note that we do not select larger values for the negative samples and down-sampling threshold because the training epoch is set to be 10 – it would require more epochs to stabilize the loss when there are more samples.

Furthermore, different studies show that performance can vary by different embedding methods [46, 48]. Thus, we also train BioConceptVec using skip-gram, GloVe and fastText, using the same default setups. We make all of the four versions of BioConceptVec (cbow, skip-gram, GloVe and fastText) publicly available so that users can experiment and choose between the models for their tasks.

To ensure a fair comparison, the evaluation datasets described below contain only concepts shared among these baseline methods and BioConceptVec. We also measured the coverage of concepts using human genes as an example.

## Intrinsic evaluations

### Identifying related genes based on drug-gene and gene-gene interactions

We posit that concept embeddings should give higher similarity to related concepts than to unrelated concepts. The intrinsic evaluations in our study quantify the effectiveness of concept embeddings in terms of identifying related genes. We concentrate on genes because genes are a central focus of biological studies; the interactions between genes (or genes and other biological concepts) are essential for understanding the structures and functions of a cell [49, 50]. In addition, biological studies over the decades have collected related genes from different perspectives, such as those based on expression signatures, pathways, and gene ontologies (GO). These collected related genes can be used as a gold standard for our intrinsic evaluations. In contrast, other biological concepts, such as diseases and mutations, are somewhat difficult to define in regard to the notion of relatedness systematically. We considered related gene pairs based on drug-gene interactions and gene-gene interactions, as explained below.

### Evaluation dataset construction and evaluation metrics

We adopted six datasets for creating evaluation datasets. The detailed statistics of these datasets are summarized in Table 3. The relatedness of genes was modeled from two broad categories. The first was based on the relationships between genes and other bio-concepts, and the second was based on the relationships among genes.

For the first category, we used the Comparative Toxicogenomics Database (CTD) [51], which captures drug-gene interactions. For each drug, we consider the genes that interact with the same drug as a related set and randomly select the same number of genes that do not interact with the drug as an unrelated set. A related and unrelated set together form a group. Ideally, concept embeddings should have significantly higher similarity for the related sets than the unrelated sets for each group.

For the second category, we used five gene sets (C1–C5) of MSigDB [52]. MSigDB captures related genes using different perspectives, and each gene set is generated from a distinct perspective. For example, MSigDB C1 is generated based on human chromosomes, and MSigDB C5 is generated based on GO. The strategy of creating related and unrelated sets is the same as above. For example, in terms of MSigDB C5, the genes that share the same GO term are considered a related set, and the same number of genes that do not share that GO term are randomly generated as an unrelated set.

We computed the similarity of a set by averaging the cosine similarity of all of the pairs in the set, using concept embeddings. Cosine similarity is the most popular similarity measure used by embeddings [29]. Importantly, different embeddings may report different cosine similarities for same pairs, and the range of cosine similarities also may be different, which is strictly inevitable [53]. To reduce the biases, for each embedding, we first applied $Z$-score standardization to the cosine similarities of all of the pairs and then used Min-Max normalization to transform the range to [0, 1].

We used the similarity score difference between related sets and unrelated sets at group level as the final evaluation metric. As noted, a more effective concept embedding should have a greater similarity score difference between the related set and the unrelated set for a group. For computational efficiency, we restricted the maximum number of genes in a set to be 100, i.e., a group has, at most, 200 genes in total. Note that MSigDB has other gene sets, such as C6 and C7. We did not use them because the number is fewer than 100 in shared genes. Collectively, our intrinsic evaluation datasets contain over 13,000 genes and over 17 million instances across six datasets.

## Extrinsic evaluations

We further evaluated the utility of BioConceptVec in two downstream applications: protein-protein interaction (PPI) prediction on the STRING database [33] and drug-drug interaction (DDI) classification on biomedical literature [54].

## Protein-protein interaction prediction on the STRING database

Analyzing functional interactions between proteins, which facilitates the understanding of the cellular processing and characterization, is a routine task in molecular systems biology [55]. The STRING database is one of the most comprehensive data resources that integrate, score, and analyze publicly available PPIs [33]. To date, it consists of over 3 billion PPIs from ~25 million proteins (https://string-db.org/). The PPIs in the STRING database are scored by accumulating a wide range of evidence, such as measurements from biological experiments, co-expressions, and gene co-occurrences.

Existing studies have used STRING for training and testing machine learning models for PPI prediction [56, 57]. In a recent study, for example, Smaili et al. constructed two PPI datasets for human proteins: (1) PPIs based on combined scores, i.e., the score calculated from multiple sources (including results from the biomedical literature and many others, such as gene co-expressions, biological experiments and pathways), which we refer to as the *combined-score*, and (2) PPIs that have the experimental score over 700, i.e., the score is based only on biological experiments and is greater than 700, which we refer to as the *experimental-700*. The study considered these PPIs as positive instances and randomly generated the same number of negative instances. Smaili et al. split the datasets into the training and testing datasets, accounting for 70% and 30% of the total number of PPIs, respectively. They further developed a deep learning model by taking the vector representations of the two proteins as inputs and predicting whether the proteins have interactions. The deep learning model was an artificial neural network (ANN) that had two hidden layers [57]. Using the same model, the study tested different vector representations and reported Area Under the Curve (AUC) accordingly.

We followed this study [57] for creating the datasets and implementing the reported ANN model. Table 4 provides a summary of the statistics of the datasets. The combined-score dataset covers all of the 13,802 proteins that are shared by concept embeddings and STRING databases. In comparison, the previous study sampled only 1,800 proteins. We also implemented a 2-layer ANN. The details of the hyperparameters are summarized in Table in S2 Table. In keeping with the previous study, the model and hyperparameters are identical when testing different concept embeddings. The Precision, Recall, F1-score, and AUC are reported.

## Drug-drug interaction extraction on biomedical literature

We also examined the usefulness of concept embeddings in a text-mining task. Specifically we evaluated the performance of concept embeddings on the SemEval 2013: Task 9 DDI extraction corpus [54] for DDI classification. This dataset consists of over 1,000 documents from the DrugBank database [58] and PubMed abstracts and ~5,000 DDIs manually annotated by two senior pharmacists, serving as a gold standard dataset for relation extraction by the community [59].

In this task, the input is a sentence that contains a pair of drugs. If the pair of drugs represents a true DDI, the model needs to output the DDI type; otherwise, the model needs to indicate the pair is not a true DDI [54]. The annotators classified a DDI into one of four types: advice, effect, mechanism, and int (the interaction occurs, but its type cannot be classified) [59]. We used the official training and testing datasets. The statistics of the datasets are summarized in Table 5. This is a multi-class classification problem (i.e., 5 classes: 4 DDI types and a negative class indicating a pair is not a DDI), and the organizers used the F1-score to measure the multi-class performance of true DDIs (i.e., without considering the negative cases). We followed the same evaluation procedure.

We implemented a simple averaged sentence embedding neural network model (SEN) for DDI classification. Fig 3 illustrates the architecture of SEN. For an input sentence, it first uses word embedding to map the vectors of each word in the sentence (Embedding Layer in Fig 3). We used the recent context-based word embedding ELMo in the Embedding Layer [60], which was shown to be superior to common word embeddings in relation extraction tasks [61]. Then it averages all of the word vectors to obtain the sentence vectors (Averaged Layer), followed by dense layers (the hidden layers used in the ANN above). Finally, it outputs class probabilities. The details of the hyperparameters of SEN are summarized in Table in S3 Table. SEN has been used widely as a baseline model in sentence-related applications [34]. We hypothesized that adding the vector representations of the drugs mentioned in the sentences will increase the classification performance. We used PubTator to map the drug mentions into concept identifiers. Thus, similar to PPI prediction, we used the same model and tested different concept embeddings. The Precision, Recall, and F-1 score are reported.

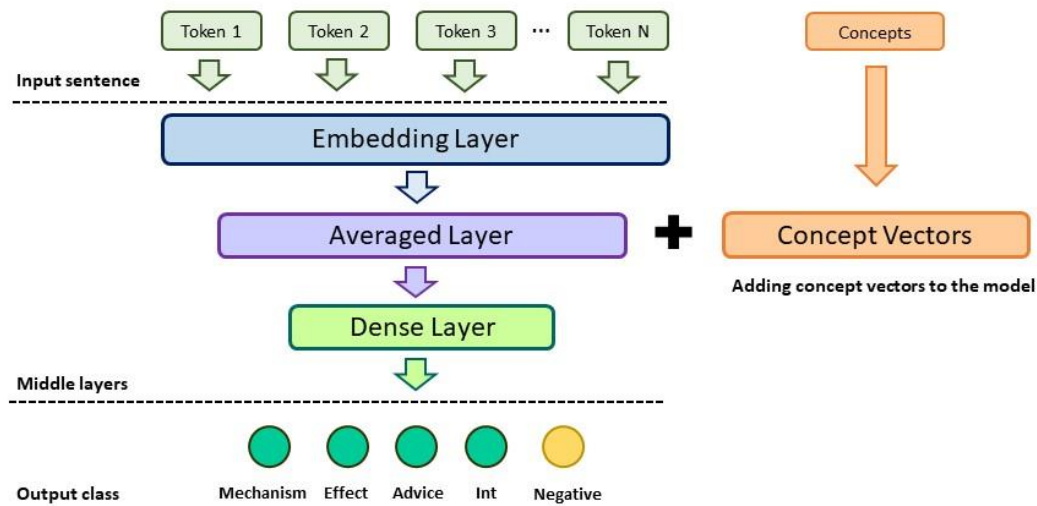

Fig 3. The architecture of the model used for DDI extraction.

# Results and Discussions

## Number of shared human genes in BioConceptVec and other public embeddings

Fig 4 shows the number of human genes with computed embeddings in each method. We compared all of the publicly available concept embeddings shown in Table 1. There are two embeddings provided by Choi et al. (https://github.com/clinicalml/embeddings). We used the version from *stanford_cuis_svd_300.txt.gz* because it contains more concepts and also more human genes than the other one. Fig 4 illustrates that BioConceptVec contains more human genes than other publicly available concept embeddings. Specifically, it covers about 3,000 more human genes than does the second highest embedding method (Yu et al). In total, these four embeddings cover 18,881 human genes, ~98% of which can be found in BioConceptVec. We manually examined the genes that were missing in BioConceptVec and found that most of them only occurred once. We also found that these genes occur more frequently in PMC full-text articles; we plan to integrate both PubMed abstracts and PMC full-text articles for training concept embeddings in the future.

Notably, the embeddings from Beam et al. and Choi et al, were primarily trained on EHR, and these embeddings are designed mainly for clinical applications. Hence, they only cover a small number of gene and protein concepts. This comparison thus further illustrates that the biomedical literature contains significantly different bio-concepts from clinical notes.

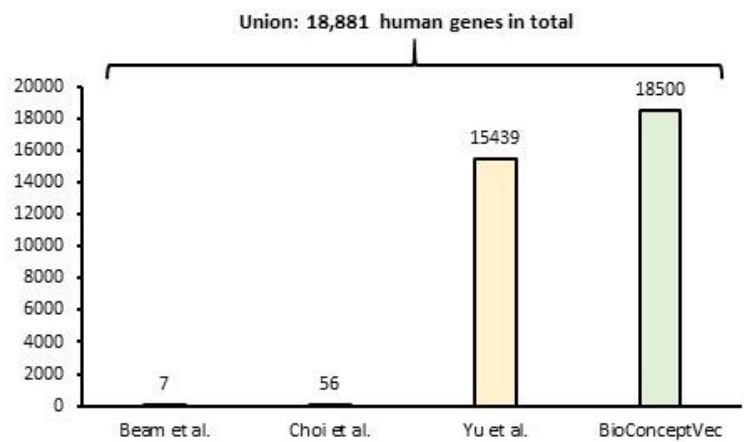

Fig 4. Gene coverage results in terms of human genes. The number of human genes in different embeddings is shown individually. In total, these four embeddings consist of 18,881 human genes. Note that the embeddings from Beam et al. and Choi et al. were mainly trained on EHR. The results mainly aim to demonstrate that biomedical literature and EHR contain significantly different concepts.

## Intrinsic evaluation results

Fig 5 and Fig 6 show the intrinsic evaluation results on the six evaluation datasets. As noted, the average group similarity difference (%) is used as the evaluation metric. A more effective concept embedding should have higher similarity difference between the positive set and the negative set of a group. Using the same embedding training method and the same hyperparameters, the results in Fig 5 show that the performance of BioConceptVec (cbow) is consistently higher (an average of 4 percentage points) than that of Yu et al. on the six datasets. The differences are even more remarked when compared to the average word embedding (an average of 7 percentage points). In addition, the results also show that BioConceptVec (cbow) achieves consistently better performance than that of baseline approaches with different hyperparameters. Collectively, these results suggest the positive impact of our selected NER methods.

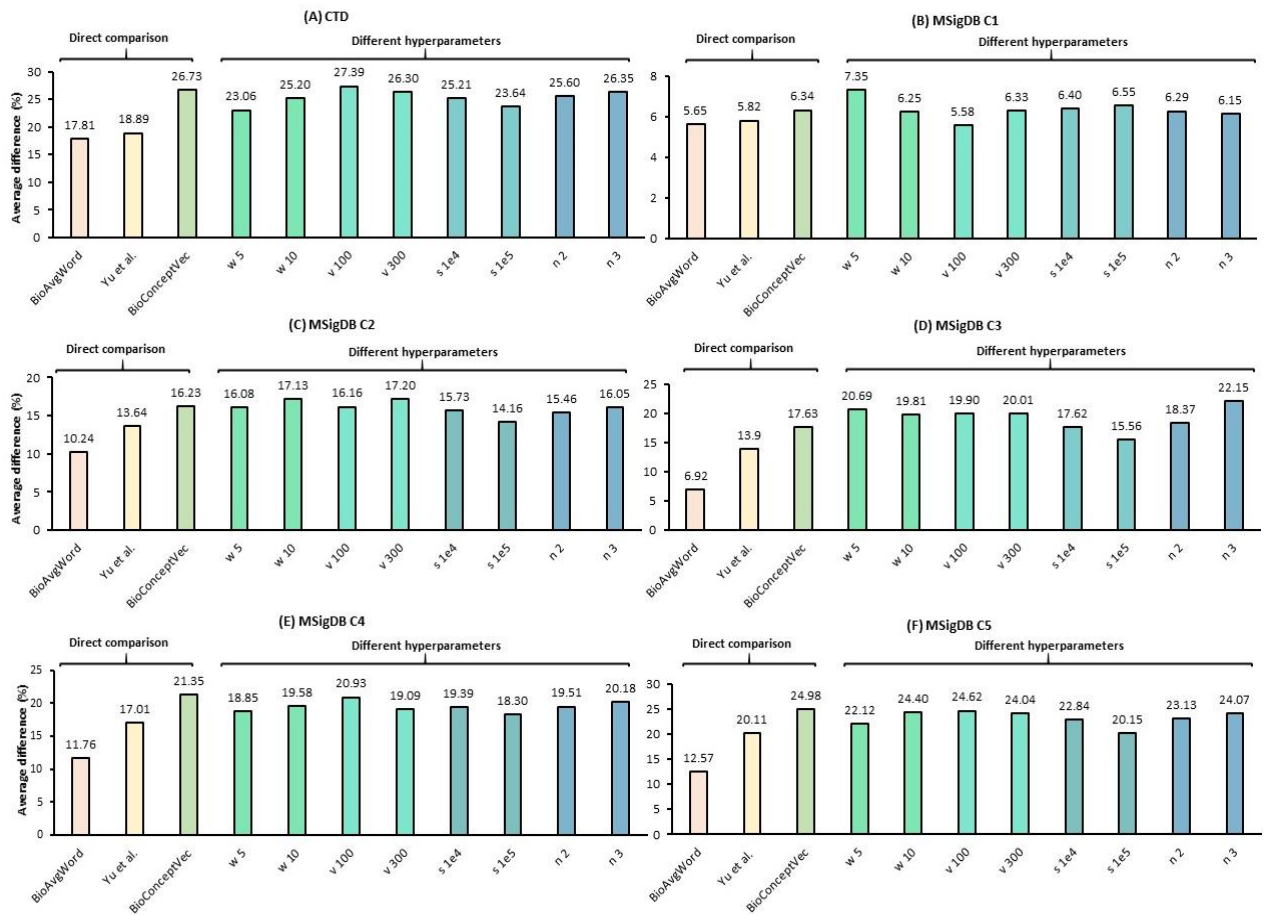

Fig 5. The intrinsic evaluation results in terms of average group similarity difference (%) for the six evaluation datasets. *Direct comparison* shows the results of BioConceptVec (cbow) using identical hyperparameters as the baselines. The baselines were also trained using cbow. *Different hyperparameters* shows the results of BioConceptVec (cbow) using different hyperparameters (provided in Table 2): w, v, s, and n stand for window size, vector dimension, sampling threshold, and negative samples, respectively.

In Fig 6, we report the effect of different embedding methods. As shown, there is no one-size-fits-all method that always achieves the best performance across all of the datasets. For instance, BioConceptVec (cbow) had the best performance on the CTD dataset, whereas BioConceptVec (GloVe) had the highest score on the MSigDB C1 dataset. This is consistent with the findings in the previous literature on embedding comparison [46, 48]. Hence, it is necessary to make embeddings trained with different methods publicly available.

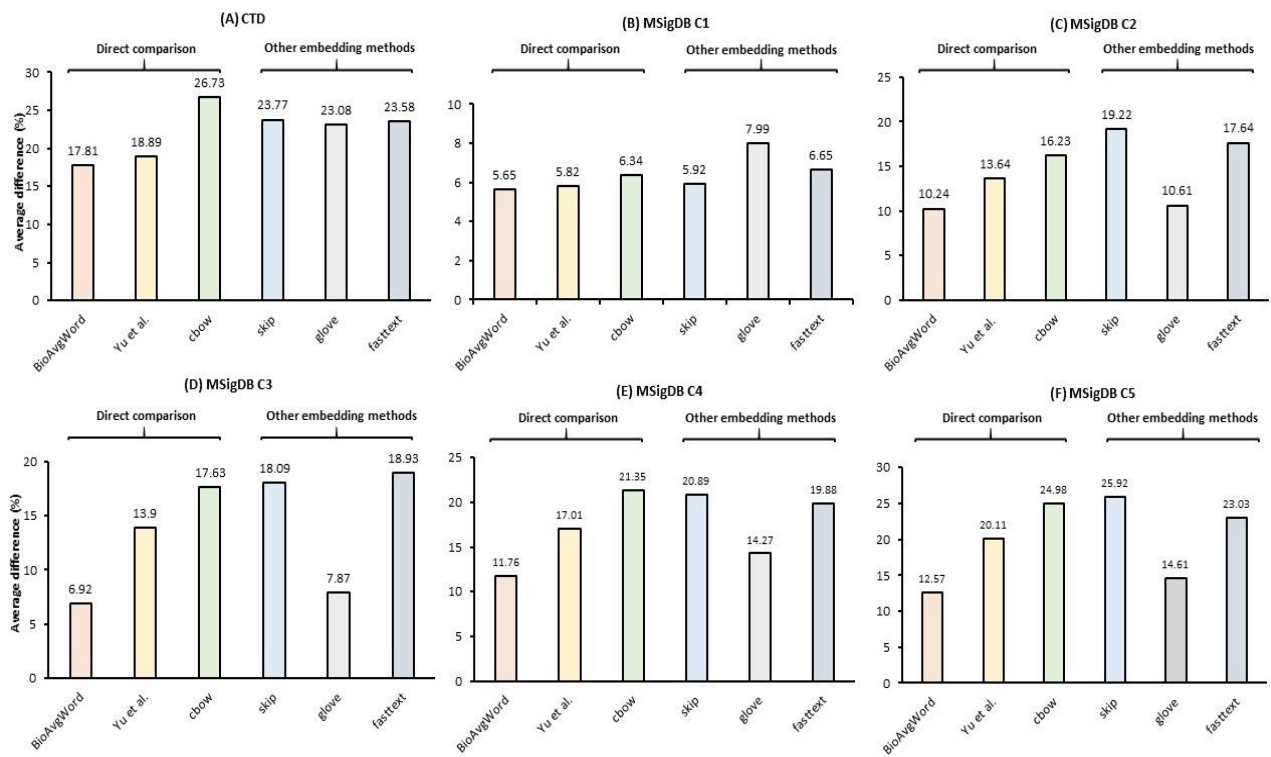

Fig 6. The intrinsic evaluation results for BioConceptVec from different embedding methods (cbow, skip-gram, Glove, and fastText). The embeddings were trained using the same default parameters. Direct comparison: the results of baseline embeddings and BioConceptVec trained using cbow.

## Extrinsic evaluation results

### Protein-protein interaction predictions on STRING database

Table 6 illustrates the classification results of PPI predictions on the STRING database. The direct comparison results show that BioConceptVec (cbow) has better performance than the baseline approaches – achieving the highest F1 score and AUC on both datasets. The results of BioConceptVec (cbow) with different hyperparameters is summarized in Table in S4 Table, which further demonstrate that its performance was consistent overall. When comparing BioConceptVec trained using different methods, BioConceptVec (fastText) had the best overall performance for this task, although the performance of BioConceptVec (cbow) and BioConceptVec (skip) are very close. Note that we were unable to directly compare with the previous study [57] because the proposed embedding is not publicly available. Also as noted, the performance of the study was measured on ~1,800 proteins, whereas our datasets contain ~13,000 proteins.

## Drug-drug interaction extraction results

Table 7 demonstrates the evaluation results on DDI extraction. We ran the model 5 times with different random seeds and then calculated the average performance [62]. The state-of-the-art (SOTA) model by Zhang and colleagues achieved an F1-score of 0.73 on this dataset [63]. Their model uses an LSTM as an encoder with an attention mechanism and outperforms other feature-based, kernel-based, and neural networks-based methods. We found that, compared with the SOTA model, the SEN model had a slightly better classification performance on advice, effect, and mechanism relation types but had a dramatically lower performance on int relation where a DDI cannot be classified into a specific type.

We also measured the performance of SEN by adding concept vectors. The direct comparison results show that BioConceptVec has better performance than the baseline approaches. Adding BioConceptVec improves the F1-score significantly and BioConceptVec (cbow) appears to be the most effective in this task. The results of BioConceptVec (cbow) using different hyperparameters are summarized in Table in S5 Table. It also shows that the performance is consistent.

We further qualitatively analyzed the errors by comparing the results of the SEN model with and without BioConceptVec. We found that the SEN model failed to classify challenging cases in which the definitions of relation types are somewhat similar. For example, the sentence, "Zidovudine competitively inhibits the intracellular phosphorylation of stavudine," contains the relation "zidovudine-stavudine." The annotator classified it as the effect type, but the SEN model wrongly classified it as the mechanism type. According to the annotation guidelines, both effect and mechanism types can describe pharmacological effects. The effect type, however, focuses on the change of the effect, whereas the mechanism type focuses on the underlying reason for the change. For this case, inhibiting the intracellular phosphorylation describes the change rather than the mechanism. There are ~20 similar erroneous cases for which the SEN model only mixed the effect type with the mechanism type. Adding BioConceptVec (cbow) to the SEN model correctly classified all of them. This is likely due to the fact that BioConceptVec provides additional information learnt from the entire PubMed abstracts, making the classification of the two related types easier as a result. Collectively, the results confirm the hypothesis that adding concept representatives improves the performance of downstream deep learning models and suggests that BioConceptVec has the potential to facilitate the development of deep learning models in the biomedical domain.

In this work, we propose BioConceptVec, concept embeddings that focus on primary biological concepts mentioned in the biomedical literature. We employed SOTA biological NER tools and trained four concept embeddings on the full collection of ~30 million PubMed abstracts. We evaluated the effectiveness of BioConceptVec in intrinsic and extrinsic settings, consisting of ~25 million instances in total. The results demonstrate that BioConceptVec consistently achieves the best performance in multiple datasets and in a range of applications. We hope that it can facilitate the development of deep learning models in biomedical research. In the future, we plan to leverage both PubMed abstracts and PMC full-text articles for training BioConceptVec.

This study focused on the evaluation on human genes because there are rich resources readily available for serving as a gold standard. We plan to evaluate BioConceptVec embeddings on different concept types in the future. Also, the quality of our concept embeddings is dependent on the accuracy of the NER tools. Improving NER tools such as PubTator would help enhance the quality of BioConceptVec. Finally, in this work, we did not apply retro-fitting, which is a fine-tuning step to further optimize the embeddings based on specific tasks with gold standard labels. For example, one of the most common retro-fitting procedures is to optimize the performance of the generated embeddings on identifying synonyms and acronyms. We did not employ it because such datasets are very limited for biomedical concepts. We plan to develop related datasets and apply the approach to further enhance BioConceptVec.

# Acknowledgments

This research was supported by the NIH Intramural Research Program, National Library of Medicine. The authors thank Dr. Alexis Allot and Dr. Robert Leaman for helpful discussions. We also thank Dr W. John Wilbur for proofreading the manuscript.

# References

1.      Singhal A, Leaman R, Catlett N, Lemberger T, McEntyre J, Polson S, et al. Pressing needs of biomedical text mining in biocuration and beyond: opportunities and challenges. Database. 2016.

2.      Lu Z, Hirschman L. Biocuration workflows and text mining: overview of the BioCreative 2012 Workshop Track II. Database. 2012.

3.      Henry S, McInnes BT. Literature based discovery: models, methods, and trends. Journal of biomedical informatics. 2017;74:20-32.

4.      Ningthoujam D, Yadav S, Bhattacharyya P, Ekbal A. Relation extraction between the clinical entities based on the shortest dependency path based LSTM. arXiv preprint arXiv:190309941. 2019.

5.      Zheng JG, Howsmon D, Zhang B, Hahn J, McGuinness D, Hendler J, et al. Entity linking for biomedical literature. BMC medical informatics and decision making. 2015;15(1):S4.

6.      Doğan RI, Kim S, Chatr-aryamontri A, Wei C-H, Comeau DC, Antunes R, et al. Overview of the BioCreative VI Precision Medicine Track: mining protein interactions and mutations for precision medicine. Database: the journal of biological databases and curation. 2019.

7.      Park J, Kim K, Hwang W, Lee D. Concept Embedding to Measure Semantic Relatedness for Biomedical Information Ontologies. Journal of Biomedical Informatics. 2019:103182.

8.      Xiang Y, Xu J, Si Y, Li Z, Rasmy L, Zhou Y, et al. Time-sensitive clinical concept embeddings learned from large electronic health records. BMC medical informatics and decision making. 2019;19(2):58.

9.      Beam AL, Kompa B, Fried I, Palmer NP, Shi X, Cai T, et al. Clinical Concept Embeddings Learned from Massive Sources of Multimodal Medical Data. arXiv preprint arXiv:180401486. 2018.

10.     Choi Y, Chiu CY-I, Sontag D. Learning low-dimensional representations of medical concepts. AMIA Summits on Translational Science Proceedings. 2016;2016:41.

11.     Ma Y, Cambria E. Concept-Based Embeddings for Natural Language Processing. arXiv preprint arXiv:180705519. 2018.

12.     Erk K. Vector space models of word meaning and phrase meaning: A survey. Language and Linguistics Compass. 2012;6(10):635-53.

13.     Li Y, Yang T. Word embedding for understanding natural language: a survey.  Guide to Big Data Applications: Springer; 2018. p. 83-104.

14.     Mikolov T, Sutskever I, Chen K, Corrado GS, Dean J, editors. Distributed representations of words and phrases and their compositionality. Advances in neural information processing systems; 2013.

15.     Pennington J, Socher R, Manning C, editors. Glove: Global vectors for word representation. Proceedings of the 2014 conference on empirical methods in natural language processing (EMNLP); 2014.

16.     Mikolov T, Grave E, Bojanowski P, Puhrsch C, Joulin A. Advances in pre-training distributed word representations. arXiv preprint arXiv:171209405. 2017.

17.     Aggarwal CC, Zhai C. Mining text data: Springer Science & Business Media; 2012.

18.     Lee K, Famiglietti ML, McMahon A, Wei C-H, MacArthur JAL, Poux S, et al. Scaling up data curation using deep learning: An application to literature triage in genomic variation resources. PLoS computational biology. 2018;14(8):e1006390.

19.     Allot A, Chen Q, Kim S, Vera Alvarez R, Comeau DC, Wilbur WJ, et al. LitSense: making sense of biomedical literature at sentence level. Nucleic acids research. 2019.

20.     Dimitriadis D, Tsoumakas G. Word embeddings and external resources for answer processing in biomedical factoid question answering. Journal of biomedical informatics. 2019;92:103118.

21.     Wei C-H, Lee K, Leaman R, Lu Z, editors. Biomedical Mention Disambiguation using a Deep Learning Approach. Proceedings of the 10th ACM International Conference on Bioinformatics, Computational Biology and Health Informatics; 2019: ACM.

22.     Du J, Jia P, Dai Y, Tao C, Zhao Z, Zhi D. Gene2vec: distributed representation of genes based on co-expression. BMC genomics. 2019;20(1):82.

23.     Bodenreider O. The unified medical language system (UMLS): integrating biomedical terminology. Nucleic acids research. 2004;32(suppl_1):D267-D70.

24.     Hassanzadeh H, Nguyen A, Koopman B, editors. Evaluation of medical concept annotation systems on clinical records. Proceedings of the Australasian Language Technology Association Workshop 2016; 2016.

25.     Lin Y-C, Christen V, Groß A, Cardoso SD, Pruski C, Da Silveira M, et al., editors. Evaluating and improving annotation tools for medical forms. International Conference on Data Integration in the Life Sciences; 2017: Springer.

26.     Reátegui R, Ratté S. Comparison of MetaMap and cTAKES for entity extraction in clinical notes. BMC medical informatics and decision making. 2018;18(3):74.

27.     Suominen H, Zhou L, Hanlen L, Ferraro G. Benchmarking clinical speech recognition and information extraction: new data, methods, and evaluations. JMIR medical informatics. 2015;3(2):e19.

28.     Pradhan S, Elhadad N, South BR, Martinez D, Christensen L, Vogel A, et al. Evaluating the state of the art in disorder recognition and normalization of the clinical narrative. Journal of the American Medical Informatics Association. 2014;22(1):143-54.

29.     Schnabel T, Labutov I, Mimno D, Joachims T, editors. Evaluation methods for unsupervised word embeddings. Proceedings of the 2015 Conference on Empirical Methods in Natural Language Processing; 2015.

30.     Choi E, Bahadori MT, Sun J, Kulas J, Schuetz A, Stewart W, editors. Retain: An interpretable predictive model for healthcare using reverse time attention mechanism. Advances in Neural Information Processing Systems; 2016.

31.     Pakhomov S, McInnes B, Adam T, Liu Y, Pedersen T, Melton GB, editors. Semantic similarity and relatedness between clinical terms: an experimental study. AMIA annual symposium proceedings; 2010: American Medical Informatics Association.

32.     Wei C-H, Kao H-Y, Lu Z. PubTator: a web-based text mining tool for assisting biocuration. Nucleic acids research. 2013;41(W1):W518-W22.

33.     Szklarczyk D, Gable AL, Lyon D, Junge A, Wyder S, Huerta-Cepas J, et al. STRING v11: protein–protein association networks with increased coverage, supporting functional discovery in genome-wide experimental datasets. Nucleic acids research. 2018;47(D1):D607-D13.

34.     Chen Q, Peng Y, Lu Z, editors. BioSentVec: creating sentence embeddings for biomedical texts. 2019 IEEE International Conference on Healthcare Informatics (ICHI); 2019: IEEE.

35.     Leaman R, Lu Z. TaggerOne: joint named entity recognition and normalization with semi-Markov Models. Bioinformatics. 2016;32(18):2839-46. Epub 2016/06/11. doi: 10.1093/bioinformatics/btw343. PubMed PMID: 27283952; PubMed Central PMCID: PMCPMC5018376.

36.     Wei C-H, Kao H-Y, Lu Z. GNormPlus: an integrative approach for tagging genes, gene families, and protein domains. BioMed research international. 2015;2015.

37.     Wei C-H, Phan L, Feltz J, Maiti R, Hefferon T, Lu Z. tmVar 2.0: integrating genomic variant information from literature with dbSNP and ClinVar for precision medicine. Bioinformatics. 2017;34(1):80-7.

38.     Wei CH, Kao HY, Lu Z. SR4GN: a species recognition software tool for gene normalization. PLoS One. 2012;7(6):e38460. Epub 2012/06/09. doi: 10.1371/journal.pone.0038460.

39.     Wei C-H, Allot A, Leaman R, Lu Z. PubTator central: automated concept annotation for biomedical full text articles. Nucleic acids research. 2019.

40.     Wang Y, Wang J, Lin H, Tang X, Zhang S, Li L. Bidirectional long short-term memory with CRF for detecting biomedical event trigger in FastText semantic space. BMC bioinformatics. 2018;19(20):507.

41.     Jin D, Szolovits P. Hierarchical Neural Networks for Sequential Sentence Classification in Medical Scientific Abstracts. arXiv preprint arXiv:180806161. 2018.

42.     Bojanowski P, Grave E, Joulin A, Mikolov T. Enriching word vectors with subword information. arXiv preprint arXiv:160704606. 2016.

43.     Yu Z, Wallace BC, Johnson T, Cohen T. Retrofitting concept vector representations of medical concepts to improve estimates of semantic similarity and relatedness. arXiv preprint arXiv:170907357. 2017.

44.     Chen Q, Du J, Kim S, Wilbur WJ, Lu Z. Deep learning with sentence embeddings pre-trained on biomedical corpora improves the performance of finding similar sentences in electronic medical records. arXiv preprint arXiv:190903044. 2019.

45.     Jang M, Kang P. Paraphrase Thought: Sentence Embedding Module Imitating Human Language Recognition. arXiv preprint arXiv:180805505. 2018.

46.     Chiu B, Crichton G, Korhonen A, Pyysalo S, editors. How to train good word embeddings for biomedical NLP. Proceedings of the 15th Workshop on Biomedical Natural Language Processing; 2016.

47.     De Vine L, Zuccon G, Koopman B, Sitbon L, Bruza P, editors. Medical semantic similarity with a neural language model. Proceedings of the 23rd ACM international conference on conference on information and knowledge management; 2014: ACM.

48.     Wang Y, Liu S, Afzal N, Rastegar-Mojarad M, Wang L, Shen F, et al. A comparison of word embeddings for the biomedical natural language processing. Journal of biomedical informatics. 2018.

49.     Barabasi A-L, Oltvai ZN. Network biology: understanding the cell's functional organization. Nature reviews genetics. 2004;5(2):101.

50.     Hartwell LH, Hopfield JJ, Leibler S, Murray AW. From molecular to modular cell biology. Nature. 1999;402(6761supp):C47.

51.     Davis AP, Grondin CJ, Johnson RJ, Sciaky D, McMorran R, Wiegers J, et al. The comparative toxicogenomics database: Update 2019. Nucleic acids research. 2018;47(D1):D948-D54.

52.     Liberzon A, Subramanian A, Pinchback R, Thorvaldsdóttir H, Tamayo P, Mesirov JP. Molecular signatures database (MSigDB) 3.0. Bioinformatics. 2011;27(12):1739-40.

53.     Zhang Y, Chen Q, Yang Z, Lin H, Lu Z. BioWordVec, improving biomedical word embeddings with subword information and MeSH. Scientific data. 2019;6(1):52.

54.     Segura Bedmar I, Martínez P, Herrero Zazo M, editors. Semeval-2013 task 9: Extraction of drug-drug interactions from biomedical texts (ddiextraction 2013)2013: Association for Computational Linguistics.

55.     Huttlin EL, Ting L, Bruckner RJ, Gebreab F, Gygi MP, Szpyt J, et al. The BioPlex network: a systematic exploration of the human interactome. Cell. 2015;162(2):425-40.

56.     Smaili FZ, Gao X, Hoehndorf R. Onto2vec: joint vector-based representation of biological entities and their ontology-based annotations. Bioinformatics. 2018;34(13):i52-i60.

57.     Smaili FZ, Gao X, Hoehndorf R. Opa2vec: combining formal and informal content of biomedical ontologies to improve similarity-based prediction. arXiv preprint arXiv:180410922. 2018.

58.     Wishart DS, Feunang YD, Guo AC, Lo EJ, Marcu A, Grant JR, et al. DrugBank 5.0: a major update to the DrugBank database for 2018. Nucleic acids research. 2017;46(D1):D1074-D82.

59.     Herrero-Zazo M, Segura-Bedmar I, Martínez P, Declerck T. The DDI corpus: An annotated corpus with pharmacological substances and drug–drug interactions. Journal of biomedical informatics. 2013;46(5):914-20.

60.     Peters ME, Neumann M, Iyyer M, Gardner M, Clark C, Lee K, et al. Deep contextualized word representations. arXiv preprint arXiv:180205365. 2018.

61.    Chauhan G, McDermott M, Szolovits P. Reflex: Flexible Framework for Relation Extraction in Multiple Domains. arXiv preprint arXiv:190608318. 2019.

62.    Peters ME, Ammar W, Bhagavatula C, Power R. Semi-supervised sequence tagging with bidirectional language models. arXiv preprint arXiv:170500108. 2017.

63.    Zhang Y, Zheng W, Lin H, Wang J, Yang Z, Dumontier M. Drug–drug interaction extraction via hierarchical RNNs on sequence and shortest dependency paths. Bioinformatics. 2017;34(5):828-35.

64.    Choi E, Bahadori MT, Searles E, Coffey C, Thompson M, Bost J, et al., editors. Multi-layer representation learning for medical concepts. Proceedings of the 22nd ACM SIGKDD International Conference on Knowledge Discovery and Data Mining; 2016: ACM.

65.    Cai X, Gao J, Ngiam KY, Ooi BC, Zhang Y, Yuan X. Medical concept embedding with time-aware attention. arXiv preprint arXiv:180602873. 2018.

66.    Nguyen K, Ichise R, editors. Learning Effective Distributed Representation of Complex Biomedical Concepts. 2018 IEEE 18th International Conference on Bioinformatics and Bioengineering (BIBE); 2018: IEEE.

# Figure Legends

Fig 1. An overview of our study. BioConceptVec was trained on full-size PubMed abstracts, which consists of ~30 million documents. (1) We employed PubTator, which contains four NER tools, to annotate and normalize the concepts. (2) We trained four concept embeddings on the normalized corpus. (3) We conducted both intrinsic evaluations on drug-gene interactions and gene-gene interactions, and extrinsic evaluations on protein-protein interaction prediction and drug-drug interaction extraction to evaluate the effectiveness of BioConceptVec.

Fig 2. Identified bio-concept in-text and the normalized versions (one instance per type) in PubTator.

Fig 3. The architecture of the model used for DDI extraction.

Fig 4. Coverage of human genes. The number of human genes in different embeddings is shown individually. In total, these four embeddings consist of 18,881 human genes. Note that the embeddings from Beam et al. and Choi et al. were mainly trained on EHR. The results mainly aim to demonstrate that biomedical literature and EHR contain significantly different concepts.

Fig 5. The average group similarity difference (%) for the six evaluation datasets. Direct comparison shows the results of BioConceptVec (cbow) using identical hyperparameters as the baselines. The baselines were also trained using cbow. Different hyperparameters shows the results of BioConceptVec (cbow) using different hyperparameters (provided in Table 2).

Fig 6. The intrinsic evaluation results for BioConceptVec using different embedding methods (cbow, skip-gram, Glove, and fastText). The embeddings were trained using the same default parameters.

# Supporting Information Legends

S1 Table. Evaluation results of the performance of the NER tools in PubTator on the concept types targeted in our study.

S2 Table. Hyperparameters of the ANN model for the protein-protein interaction prediction.

S3 Table. Hyperparameters of the SEN model for the drug-drug interaction prediction.

S4 Table. Classification results of BioConceptVec (cbow) trained using different hyperparameters for the protein-protein interaction prediction.

S5 Table. Classification results of BioConceptVec (cbow) trained using different hyperparameters for the drug-drug interaction prediction.

# Tables

Table 1. An overview of biomedical concept embeddings trained on large-scale free-text corpora. Repository: the scope of concepts. Corpora: the training collection. Note that for EHR (electronic health records) and Claims (medical claims), the size is the number of patients, whereas for Wikipedia, PubMed (abstracts), and PMC (full-text articles), the size is the number of documents. #Concepts: the number of distinct concepts in the embedding. Method: the method for training embeddings. PCA: principle component analysis. PMI: pointwise mutual information. Intrinsic evaluation: a focus on applications that directly use the similarity between the vectors produced by word embeddings, such as word-pair similarity and relatedness. Extrinsic evaluation: a focus on downstream applications that use only word embeddings as an intermediate component. For example, the last study evaluated the effectiveness of concept embeddings for heart-failure prediction. Availability: whether the studies made the embeddings publicly available (we accessed on 04/20/2019).

| Study (year) | Repository | Corpora (size) | #Concepts | Method | Evaluation | | Availability |
|---|---|---|---|---|---|---|---|
| | | | | | Intrinsic | Extrinsic | |
| Vine et al. (2014) [47] | UMLS | EHR (<20K) PubMed (0.35M) | 52,102 | skip-gram | Concept similarity | N | N |
| Choi et al. (2016) [64] | ICD9CM | EHR (0.55M) Claims (0.85M) | 49,873 | skip-gram | Concept clustering | N | N |
| Choi et al. (2016) [10] | UMLS | EHR (20M) Claims (4M) | 22,705 | skip-gram | Concept clustering | N | Y |
| Yu at al. (2017) [43] | UMLS | PubMed (22M) | 310,403 | cbow | Concept similarity | N | Y |
| Beam et al. (2018) [9] | UMLS | EHR (60M) Claims (20M) PMC (1.7M) | 108,477 | skip-gram GloVe PCA | Concept similarity | N | Y |
| Cai at al. (2018) [65] | UMLS | EHR (2M) | 47,873 | cbow | Concept clustering | N | N |
| Nguyen at al. (2018) [66] | UMLS | Wikipedia (5M) PubMed (24M) PMC (3M) | 659,873 | cbow | Concept similarity | N | N |
| Xiang at al. (2019) [8] | UMLS | EHR (50M) | 30,348 | skip-gram PMI fastText | Concept clustering | Y | N |

Table 2. The values of hyperparameters used for training BioConceptVec. *Default values*: the default values are identical to the values selected by baseline embeddings. We used the default values to train BioConceptVec (cbow), BioConceptVec (skip-gram), BioConceptVec (GloVe) and BioConceptVec (fastText). *Other values*: we also adopted other commonly-used hyperparameter values to test the effectiveness of BioConceptVec (cbow) under different parameter settings.

| | Hyperparameter | Default values | Other values |
|---|---|---|---|
| Shared hyperparameters | Vector dimension | 200 | 100, 300 |
| | Window size | 20 | 5, 10 |
| | Negative samples | 5 | 2, 3 |
| | Down-sampling threshold | 0.001 | 0.0001, 0.00001 |
| | Minimal word occurrence | 5 | - |
| | Learning rate | 0.025 | - |
| | Training epochs | 10 | - |
| fastText-specific hyperparameters | Minimal character n-gram length | 2 | - |
| | Maximum character n-gram length | 3 | - |

Table 3. The statistics of datasets in intrinsic evaluation tasks. There are six datasets in total. #groups: the number of groups in a dataset. Each group has a related set and an unrelated set of genes based on drug-gene interactions provided by CTD or gene sets provided by MSIGDB. #distinct concepts: the total number of distinct genes in a dataset. Avg #concepts per group: the average of number of genes in a group; note that one gene may be in multiple groups.  #pairs: the total number of pairs in a dataset. Avg #pairs per group: the average of the number of pairs per group.

| Dataset | #groups | #distinct concepts | Avg #concepts per group | #pairs | Avg #pairs per group |
|---|---|---|---|---|---|
| CTD | 6383 | 14,654 | 22.39 | 2,146,482 | 358.88 |
| MSigDB datasets | | | | | |
| C1 positional gene sets | 326 | 11,709 | 63.30 | 431,254 | 1447.16 |
| C2 curated gene sets | 4,762 | 13,783 | 66.21 | 6,171,976 | 1621.21 |
| C3 motif gene sets | 836 | 9,553 | 115.63 | 910,722 | 3976.95 |
| C4 computational gene sets | 858 | 8,637 | 85.84 | 1,452,542 | 2392.99 |
| C5 GO gene sets | 5,917 | 13,627 | 62.71 | 6,697,736 | 1455.08 |
| Total | 19,082 | 14,998 | - | 17,810,712 | - |

Table 4. Statistics of the datasets for PPI prediction. #Concepts: the number of concepts in the dataset. #Training: the number of training instances; same applies to #Validation and #Testing.

| Dataset | #Concepts | #Training | #Validation | #Testing | Total |
|---|---|---|---|---|---|
| combined-score | 13,802 | 5,245,358 | 582,818 | 2,497,790 | 8,325,966 |
| experimental-700 | 13,290 | 24,684 | 2,743 | 11,755 | 39,182 |

Table 5. Statistics of the DDI extraction datasets. Mechanism, Effect, Advice, Int are four types of DDIs. Negative means that the instance does not contain a DDI.

| Class | #Training | #Testing |
|---|---|---|
| Mechanism | 1,319 | 302 |
| Effect | 1,621 | 360 |
| Advice | 826 | 221 |
| Int | 188 | 96 |
| Negative | 23,772 | 4,737 |

Table 6. Classification results of PPI predictions on the STRING database. Combined-scores: PPIs that have combined scores are considered positive cases. Experimental-700: PPIs that have experimental scores over 700 are considered positive cases. Direct comparison: the results of embeddings using the same method (cbow) and same hyperparameters. Different embedding methods: the results of BioConceptVec (skip-gram), BioConceptVec (GloVe) and BioConceptVec (fastText). The highest results of each section are marked as bold.

| | Combined-score dataset | | | | Experimental-700 dataset | | | |
|---|---|---|---|---|---|---|---|---|
| | Precision | Recall | F1 | AUC | Precision | Recall | F1 | AUC |
| **Direct comparison** | | | | | | | | |
| BioAvgWord (cbow) | 0.8195 | 0.7935 | 0.8063 | 0.8941 | 0.8851 | 0.7422 | 0.8074 | 0.9123 |
| Yu et al. (cbow) | 0.8236 | 0.8017 | 0.8125 | 0.9029 | 0.9130 | 0.7686 | 0.8346 | 0.9283 |
| BioConceptVec (cbow) | **0.8304** | **0.8025** | **0.8162** | **0.9064** | **0.9476** | **0.7981** | **0.8664** | **0.9525** |
| **Different embedding methods** | | | | | | | | |
| BioConcept (skip-gram) | 0.8279 | 0.8097 | 0.8187 | 0.9074 | **0.9201** | 0.8525 | 0.8850 | 0.9522 |
| BioConcept (GloVe) | 0.8116 | **0.8102** | 0.8109 | 0.9004 | 0.8656 | 0.8289 | 0.8468 | 0.9218 |
| BioConcept (fastText) | **0.8324** | 0.8100 | **0.8210** | **0.9099** | 0.9076 | **0.8677** | **0.8872** | **0.9556** |

Table 7. Classification results of DDI classification. SOTA: state-of-the-art. P: Precision. R: Recall. The SOTA results are extracted from [63]. Direct comparison: the results of embeddings using the same method (cbow) and same hyperparameters. Different embedding methods: the results of BioConceptVec (skip-gram), BioConceptVec (GloVe) and BioConceptVec (fastText). The highest results of each section are marked as bold.

| Model | F1-score on each relation type | | | | Overall performance | | |
|---|---|---|---|---|---|---|---|
| | Int | Advice | Effect | Mechanism | P | R | F |
| Zhang et al. (SOTA) | **0.5400** | 0.8000 | 0.7200 | 0.7400 | 0.7400 | 0.7200 | 0.7300 |
| SEN | 0.3569 | **0.8336** | **0.7978** | **0.8463** | **0.7940** | **0.7832** | **0.7776** |
| **Direct comparison** | | | | | | | |
| SEN + BioAvgWord (cbow) | 0.3150 | 0.7787 | 0.8000 | 0.8824 | 0.7883 | 0.7814 | 0.7731 |
| + Yu et al. (cbow) | 0.4285 | 0.8263 | 0.8133 | 0.8559 | 0.7948 | 0.7961 | 0.7916 |
| + BioConceptVec (cbow) | **0.5206** | **0.8423** | **0.8191** | **0.8692** | **0.8167** | **0.8161** | **0.8105** |
| **Different embedding methods** | | | | | | | |
| + BioConcept (skip-gram) | 0.4090 | **0.8164** | **0.8255** | 0.8626 | **0.8088** | 0.8025 | 0.7941 |
| + BioConcept (GloVe) | **0.4587** | 0.8100 | 0.8160 | **0.8702** | 0.8046 | **0.8029** | **0.7963** |
| + BioConcept (fastText) | 0.4382 | 0.8153 | 0.8200 | 0.8571 | 0.7999 | 0.7998 | 0.7930 |